\documentclass[letterpaper]{article} 
\usepackage{aaai2027} 
\usepackage[hyphens]{url} 
\usepackage{graphicx} 
\usepackage{natbib} 
\usepackage{caption} 
\usepackage{amsmath}
\usepackage{amssymb}
\usepackage{booktabs}

\title{An Integrated Video-AI Platform for Action-Level Microanastomosis Training and Performance Feedback}
\author{
	Yan Meng\textsuperscript{\rm 1}\corresponding,
	Daniel A. Donoho\textsuperscript{\rm 1}
}
\affiliations{
	\textsuperscript{\rm 1}Department of Neurosurgery, Children’s National Hospital\\	
	ymeng@childrensnational.org, ddonoho@childrensnational.org
}

\begin{document}
\onecolumn
\begingroup
\renewcommand{\twocolumn}[1][]{#1}
\maketitle
\endgroup
\raggedbottom
\renewcommand{\topfraction}{0.9}
\renewcommand{\bottomfraction}{0.8}
\renewcommand{\textfraction}{0.05}
\renewcommand{\floatpagefraction}{0.85}
\setcounter{topnumber}{4}
\setcounter{bottomnumber}{3}
\setcounter{totalnumber}{6}

\begin{abstract}
Developing microanastomosis skill requires repeated practice with timely, action-specific feedback, yet expert review of lengthy microscope videos does not scale to frequent or distributed training. We present an integrated video-AI platform that turns a complete simulated procedure into inspectable, interactive feedback through three connected modules. First, a proposed transformer segments the video into six surgical actions. Second, object detection and tracking localize instrument tips within each action; the resulting kinematic features and action statistics drive supervised classification of five NOMAT-aligned performance dimensions. Third, a grounded large language model (LLM) uses these structured outputs to answer user questions about the current scene, actions, motion, and predicted performance through a unified interface. In a two-site study, 17 participants completed 72 procedures comprising 576 suture placements. The action-segmentation module achieved 87.66\% accuracy and 82.86\% F1, increasing to 93.62\% and 88.32\% after workflow-aware refinement. The five performance classifiers achieved 76.0\% mean accuracy, with Cohen's $\kappa$ from 0.63 to 0.93. Although the language interface and educational benefit require prospective evaluation, these results establish the technical basis for an expert-supervised platform that can shorten review, expose the evidence behind performance estimates, and support scalable formative microsurgical training.
\end{abstract}

\section{Introduction}
Microanastomosis is a foundational neurosurgical skill that requires precise manipulation of submillimeter vessels under high magnification. Proficiency develops through deliberate practice with timely, specific feedback. Assessment typically depends on direct faculty observation or retrospective review of lengthy recordings using instruments such as the Objective Structured Assessment of Technical Skill, global rating scales, and the Neurosurgical Objective Microanastomosis Assessment Tool (NOMAT) \cite{martin1997objective,regehr1998comparing,aoun2015pilot}. This approach is difficult to sustain when learners practice repeatedly or train at geographically distributed sites, especially where specialist surgical education is limited \cite{meara2015global}.

Video-based assessment can reduce this bottleneck without requiring sensors on the instruments. Previous studies have used learned visual representations, attention, robotic kinematics, and hand-crafted motion features to distinguish surgical skill levels \cite{funke2019video,wang2018satr,kitaguchi2021development,li2022surgical,yanik2023video,goldbraikh2022video,hung2023capturing,zia2016automated,lavanchy2021automation,meng2023automatic}. Temporal convolutional networks such as MS-TCN and TeCNO have been applied to surgical workflow analysis \cite{farha2019ms,czempiel2020tecno}, while TimeSformer, ASFormer, ActionFormer, ASTCFormer, and Surgformer have advanced temporal modeling in general and surgical video \cite{bertasius2021space,yi2021asformer,zhang2022actionformer,zhang2023surgical,yang2024surgformer}. However, a procedure-level score alone provides limited guidance. Training feedback is more useful when it identifies recognizable actions and connects predictions to interpretable evidence.

We address this gap with a single training platform organized around three connected analysis modules and a shared user interface. The action-segmentation module determines what occurs and when. The object-analysis module detects and tracks surgical instruments within those action intervals, then combines kinematic features with action duration, repetition, and cumulative time to classify performance. The LLM interaction module receives these structured, video-grounded outputs and answers scene- and performance-related questions without replacing the underlying measurements. The application is intended for formative assessment in simulation and does not replace faculty judgment or determine readiness for independent clinical practice.

Our contributions are:
\begin{itemize}
    \item an integrated platform that connects video upload, action segmentation, instrument analysis, supervised performance classification, and grounded natural-language interaction in one review interface;
    \item a proposed action-segmentation model combining hierarchical temporal attention (HTA) and variance-weighted spatial self-attention (VWSSA) for fine-grained microanastomosis video;
    \item action-conditioned instrument tracking and interpretable kinematic and temporal features for five NOMAT-aligned performance classifiers; and
    \item a two-site evaluation with 17 participants, 72 procedures, 576 stitch-level clips, and a preliminary user study of the training workflow.
\end{itemize}

\par\medskip
\begin{figure}[htbp]

\centering

    \centering
    \includegraphics[width=0.82\linewidth,height=0.34\textheight,keepaspectratio]{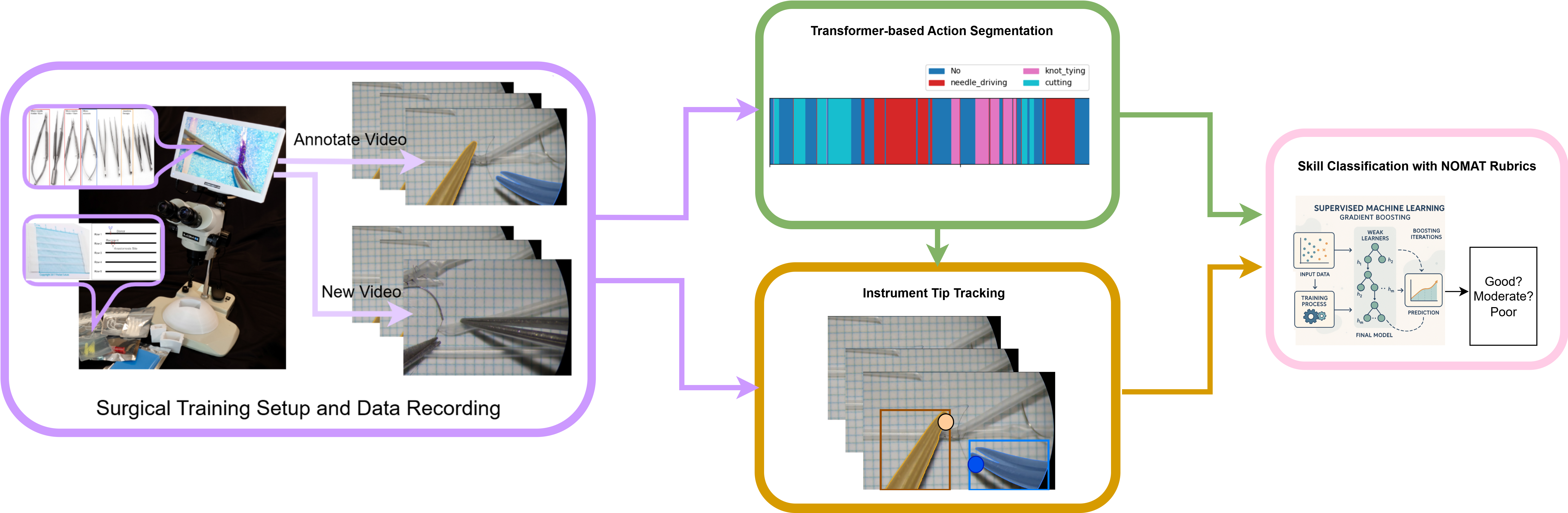}
    \caption{Integrated platform workflow. The action-segmentation module divides a microscope video into surgical actions. The object-analysis module detects and tracks instruments within each action and combines kinematics with action statistics for supervised performance classification. A unified interface exposes these outputs to users and grounds an LLM for scene- and performance-related questions.}
    \label{fig:full_framework}
\end{figure}
\par\medskip

\section{Method}
Figure~\ref{fig:full_framework} summarizes the complete platform. Its modules exchange explicit intermediate representations rather than independent predictions: a frame-level action timeline conditions instrument and feature analysis; action-specific trajectories and statistics condition supervised performance classification; and the timeline, tracked objects, feature summaries, and provisional scores form the evidence supplied to the interactive LLM. This design keeps every response traceable to a video interval or computed measurement.

\subsection{Practice Kit and Application Workflow}
\begin{figure}[htbp]

\centering

    \centering
    \includegraphics[width=0.82\linewidth,height=0.34\textheight,keepaspectratio]{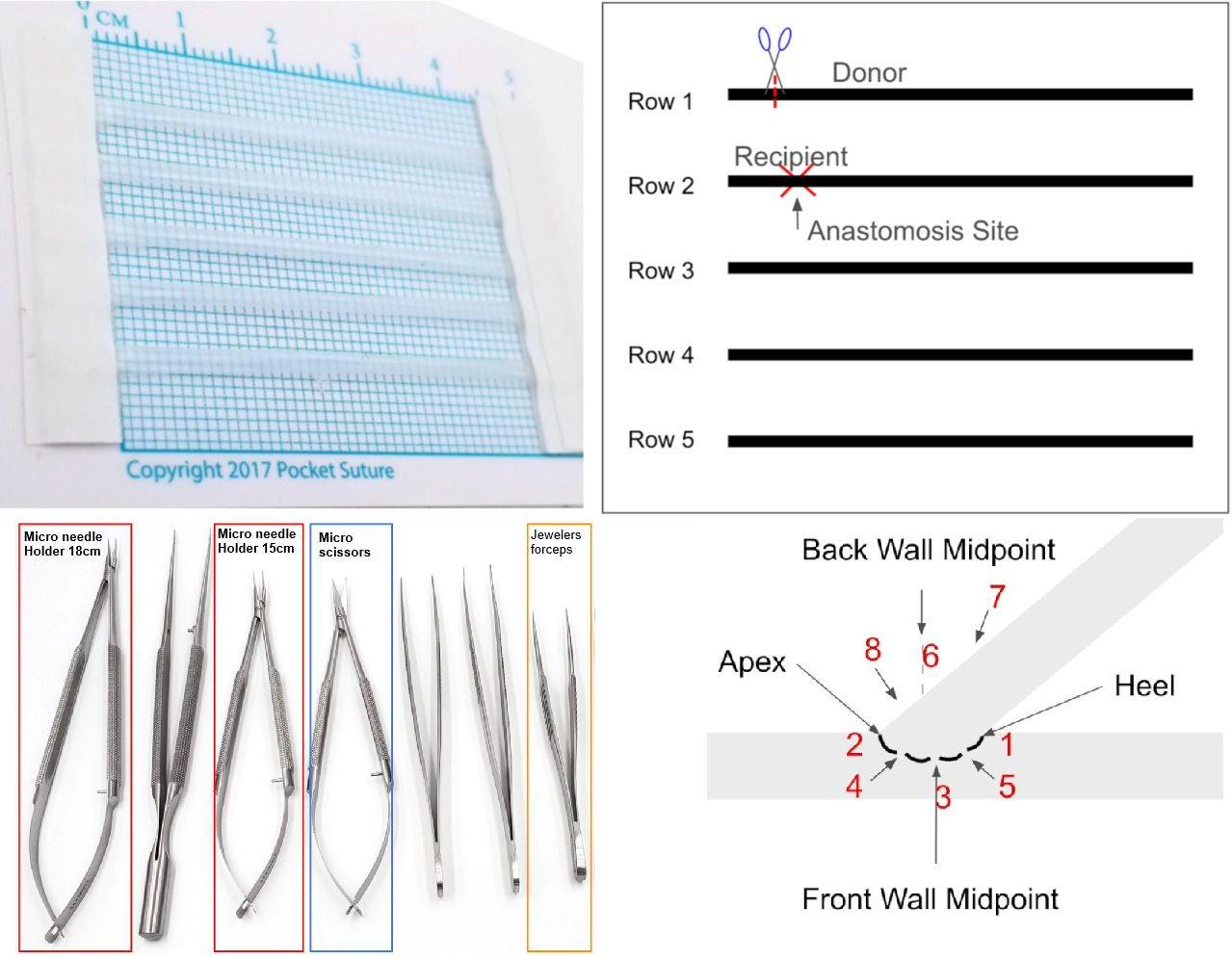}
    \caption{Synthetic microvascular practice card and standardized eight-suture placement pattern.}
    \label{fig:full_card}
\end{figure}

The kit contains a trinocular zoom stereomicroscope, high-definition camera, external monitor, synthetic microvascular cards, and a uniform instrument set. Participants prepare donor and recipient vessels and complete an end-to-side anastomosis by placing eight sutures in a standardized order (Figure~\ref{fig:full_card}). The microscope camera records the procedure without instrument-mounted sensors. A user can review the resulting video in the interface while the three modules provide synchronized action, object, performance, and conversational views.

\par\medskip
\par\medskip

\subsection{Transformer-Based Action Segmentation}
The model builds on TimeSformer \cite{bertasius2021space} and predicts vessel cutting, needle handling, needle touching the vessel, needle withdrawal, knot tying, knot cutting, and background. HTA combines global attention across the full $T$-frame input with local attention over $T/2$ and $T/4$ windows. The global branch represents procedural context, while local branches capture brief motions and action boundaries.

\begin{figure}[htbp]

\centering

    \centering
    \includegraphics[width=0.82\linewidth,height=0.34\textheight,keepaspectratio]{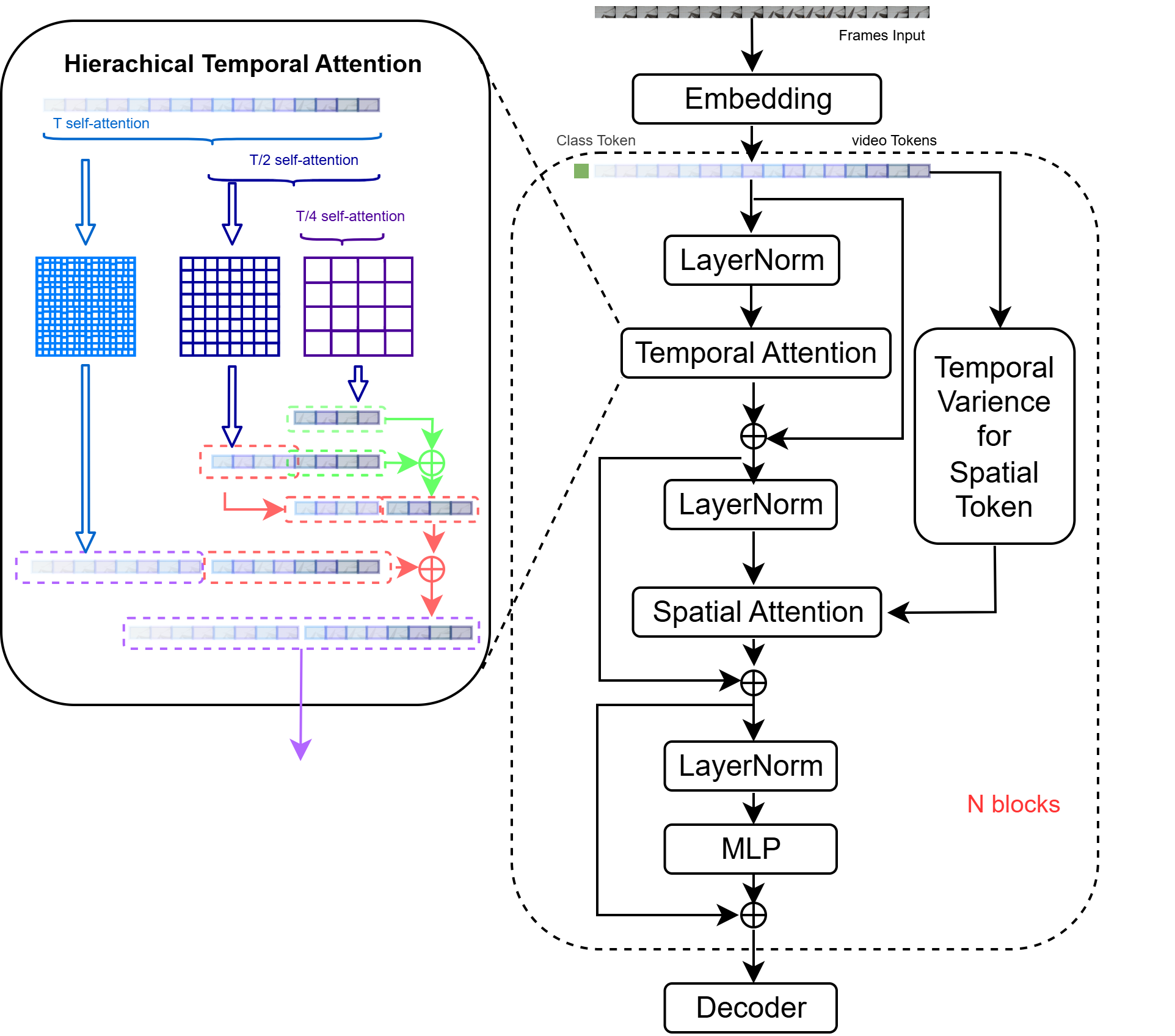}
    \caption{Action-segmentation architecture with hierarchical temporal attention and variance-weighted spatial attention.}
    \label{fig:full_transformer}
\end{figure}

VWSSA emphasizes spatial regions that change across time. For spatial token $i$, temporal variance $\sigma_i^2$ defines
\begin{equation}
w_i=\frac{\exp(\sigma_i^2)}{\sum_j\exp(\sigma_j^2)}.
\end{equation}
These weights favor moving instruments and tissue interactions over static background. Temporal smoothing removes predicted segments shorter than five frames, and an action dictionary corrects implausible transitions. Figure~\ref{fig:full_transformer} shows the architecture.

\par\medskip
\par\medskip

\subsection{Instrument Detection and Tip Tracking}
YOLO detection \cite{redmon2016you} is combined with DeepSORT tracking \cite{wojke2017simple}, following prior work on learned surgical-instrument tracking \cite{qiu2019real}. High magnification produces abrupt apparent motion, occlusion, and visually similar instruments. High-confidence detections correct drifting tracker boxes, reliable class labels are propagated along persistent identities, and identities are reassigned after short occlusions using instrument class, temporal gap, and appearance similarity \cite{meng2025ai}.

Candidate tip points are sampled on the convex hull of each instrument silhouette. Each descriptor $\mathbf d_i$ is compared with a reference tip descriptor $\mathbf d_{\mathrm{ref}}$:
\begin{equation}
\hat p=\arg\max_i
\frac{\mathbf d_{\mathrm{ref}}\cdot\mathbf d_i}
{\lVert\mathbf d_{\mathrm{ref}}\rVert\lVert\mathbf d_i\rVert}.
\end{equation}
The selected points are mapped to global frame coordinates. The trajectories yield velocity, acceleration, jerk, inter-instrument distance, relative speed, and angular displacement.

\subsection{NOMAT-Aligned Skill Classification}
A 32-dimensional vector summarizes individual-instrument and relative motion. For each action, duration, repetition count, and cumulative time are also computed. Gradient Boosting Classifiers model nonlinear relationships in these tabular features \cite{konstantinov2021interpretable}. Separate classifiers predict Poor, Moderate, or Good performance for overall instrument handling, needle-handling motion quality, knot-tying motion quality, needle-action efficiency, and knot-tying efficiency.

\subsection{Grounded LLM Interaction and User Interface}
The language module converts the outputs of the first two modules into an interactive review experience. For the selected time point or action interval, the application constructs a structured context containing the dominant action and its confidence, detected instrument identities and tip locations, recent kinematic summaries, action duration and repetition statistics, and the provisional NOMAT-aligned classifications. The LLM is instructed to answer from this context and to distinguish computed observations from general training suggestions. Users can ask what action is occurring, which instruments are visible, how the instruments are moving, what performance dimension is lower, or which source interval supports an answer.

\begin{figure}[htbp]

\centering

    \centering
    \includegraphics[width=0.82\linewidth,height=0.34\textheight,keepaspectratio]{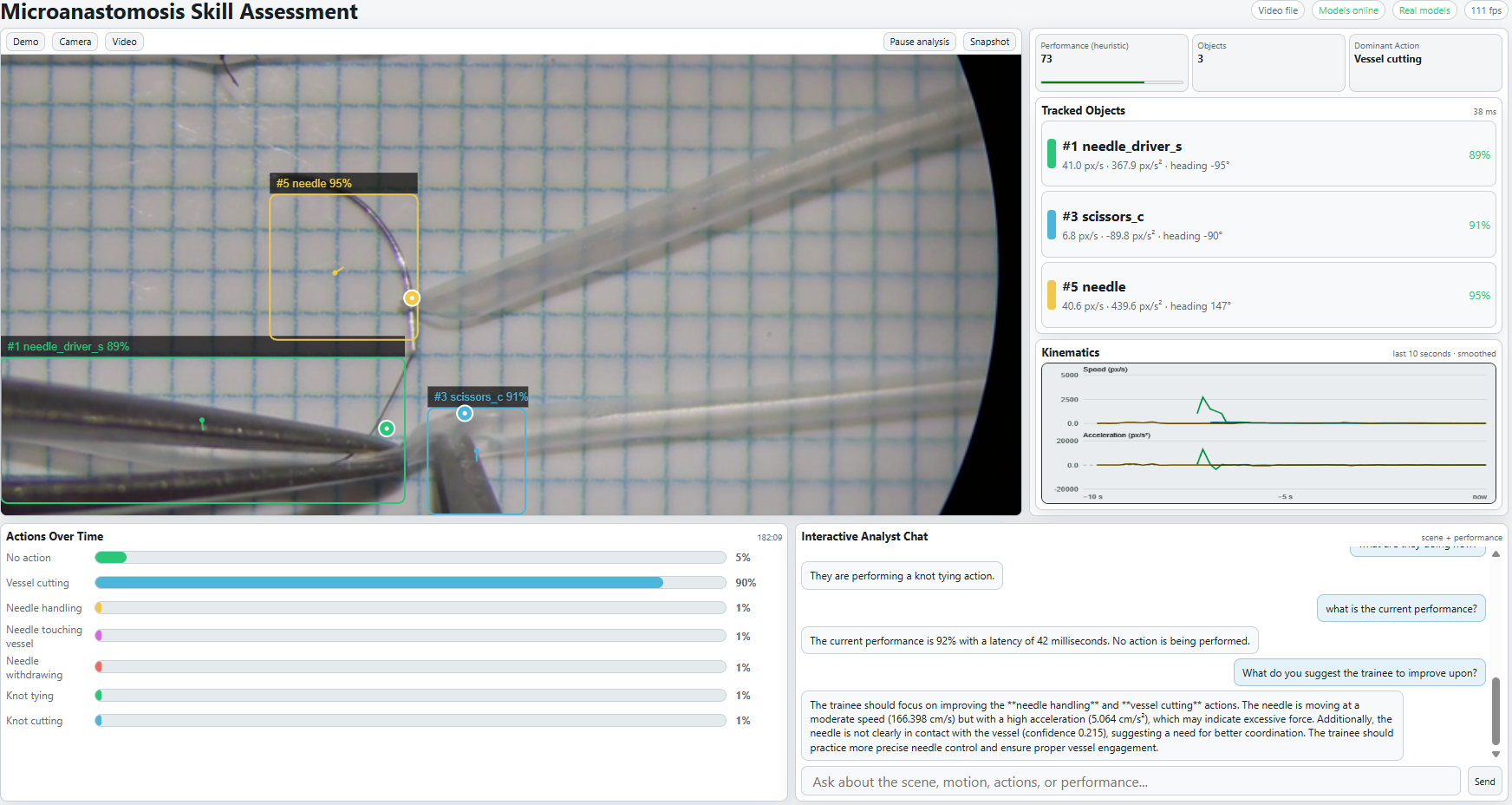}
    \caption{Unified platform interface showing the source video, instrument detections and tip tracks, action timeline, live kinematics, provisional performance, and an LLM chat grounded in scene and performance outputs.}
    \label{fig:full_ui}
\end{figure}

The LLM does not analyze raw pixels or generate the performance score. It serves as a natural-language access layer over the platform's measured outputs, preserving the action segmentation and supervised classifiers as the sources of record. When the requested information is absent or model confidence is insufficient, the interface should abstain and direct the user to the video or faculty review. Figure~\ref{fig:full_ui} shows the unified workspace.

\par\medskip
\par\medskip

\subsection{Browser Acquisition and Runtime Scheduling}
The deployed prototype, MicroVision-BYPASS, accepts prerecorded video or live camera frames in a browser and transmits JPEG frames to a FastAPI service. Detection requests are sent serially as the preceding request completes, while the 16-frame action worker updates independently once per second. Rendering runs separately from inference so a temporal update does not block the instrument overlay. The one-second schedule describes output refresh, not measured inference latency.

Persistent workers keep each neural model in its original software environment and exchange newline-delimited JSON with the orchestration service. Detection and action workers start at initialization; the local Qwen3-8B worker loads on the first question. Serialized worker calls prevent responses from becoming interleaved. This separation supports models with different update rates and dependencies without reloading checkpoints for every frame.

\subsection{Detection and Measurement in the Prototype}
The runtime uses a fine-tuned YOLO11x detector for straight and curved needle drivers, straight and curved scissors, and needle. Deep SORT preserves temporal identities. The current implementation retains at most one representative output per class, an assumption appropriate only to the standardized instrument arrangement. Multiple simultaneous tools of the same class would require a different association policy.

The earlier motion-assessment pipeline uses shape-descriptor matching to localize tips. The deployed lightweight localizer instead combines thresholding, morphology, contour geometry, motion, center-facing orientation, and temporal continuity. These are related implementations, not interchangeable validation results. Both operate inside detector boxes and map candidate positions back into image coordinates.

The dashboard's live speed and acceleration are derived from box-centroid displacement and are reported in pixels/s and pixels/s$^2$. They should be distinguished from calibrated tip trajectories used for quantitative motion assessment. Box deformation can move a centroid without corresponding tip movement, and changing magnification alters pixel distances. Meaningful cross-session comparison requires spatial calibration, consistent timestamps, trajectory stabilization, and filtering before computing acceleration or jerk. Neither image-plane acceleration nor jerk directly measures tissue force.

\section{Experiment}
\subsection{Participants and Data Collection}
The study was conducted at two collaborating sites with Institutional Review Board approval. Seventeen participants were enrolled: nine residents (52.9\%), six medical students (35.3\%), and two fellows (11.8\%). Fourteen (82.4\%) were male and three (17.6\%) were female; mean age was approximately 30 years.

Participants completed repeated procedures under the standardized protocol, producing 72 recordings with an average duration of approximately 26 minutes. Each procedure contained eight suture placements, yielding 576 stitch-level clips. Twenty procedures comprising more than 1,120 action segments were manually annotated: 15 for training, three for validation, and two for testing. The trained model segmented the remaining 52 procedures.

Two board-certified neurosurgeons independently evaluated every procedure using NOMAT and reconciled disagreements by consensus. Five-point scores were grouped into Poor, Moderate, and Good categories using thresholds of 2.5 and 3.5.

\subsection{Training and Evaluation}
The segmentation model was initialized with Kinetics-pretrained TimeSformer weights \cite{kay2017kinetics,bertasius2021space}. Videos were sampled at 10 frames per second with $T=16$. Training used 50 epochs, batch size 16, four NVIDIA T4 GPUs, AdamW with learning rate $9\times10^{-5}$, and layer-wise decay 0.75.

Segmentation was evaluated with frame accuracy and action-level precision, recall, Jaccard index, and F1 score. Comparisons included MS-TCN, Surgformer, and VideoMamba. Skill classifiers used an 80/20 train-test split with five-fold cross-validation inside the training subset. Accuracy measured agreement with consensus expert categories, and Cohen's $\kappa$ quantified agreement for each NOMAT dimension.

The present study quantitatively evaluates the action-segmentation and supervised performance-classification components and descriptively evaluates participant perceptions of the training workflow. The LLM-enabled interface is evaluated as an integrated prototype demonstration; no independent benchmark of answer correctness, grounding, latency, or usability was conducted. We therefore report no unsupported quantitative claim for the language module.

\subsection{User Study}
All 17 participants completed a baseline survey concerning experience, confidence, and video-based assessment. A follow-up survey was offered to the 11 participants who completed at least five trials; three responded. Baseline and follow-up responses were not linked, so the survey was analyzed descriptively rather than as paired individual changes.

\section{Results}
\subsection{Action Segmentation}
The proposed model achieved 87.66\% accuracy and 82.86\% F1 score and outperformed all comparison methods (Table~\ref{tab:full_segmentation}). Post-processing increased accuracy to 93.62\% and F1 to 88.32\%.

\par\medskip
\begin{table}[htbp]

\centering
\begin{tabular}{lccccc}
\toprule
Method & Accuracy & Precision & Recall & Jaccard & F1 \\
\midrule
MS-TCN & 78.64 & 75.17 & 77.64 & 60.54 & 71.44 \\
Surgformer & 82.47 & 80.82 & 79.54 & 68.60 & 75.96 \\
VideoMamba & 80.80 & 77.61 & 78.99 & 65.37 & 75.45 \\
Ours & 87.66 & 83.29 & 83.99 & 71.57 & 82.86 \\
Ours + post-processing & \textbf{93.62} & \textbf{89.32} & \textbf{88.71} & \textbf{82.22} & \textbf{88.32} \\
\bottomrule
\end{tabular}
\caption{Action-segmentation performance (\%).}
\label{tab:full_segmentation}
\end{table}
\par\medskip

\begin{figure}[htbp]

\centering

    \centering
    \includegraphics[width=0.82\linewidth,height=0.34\textheight,keepaspectratio]{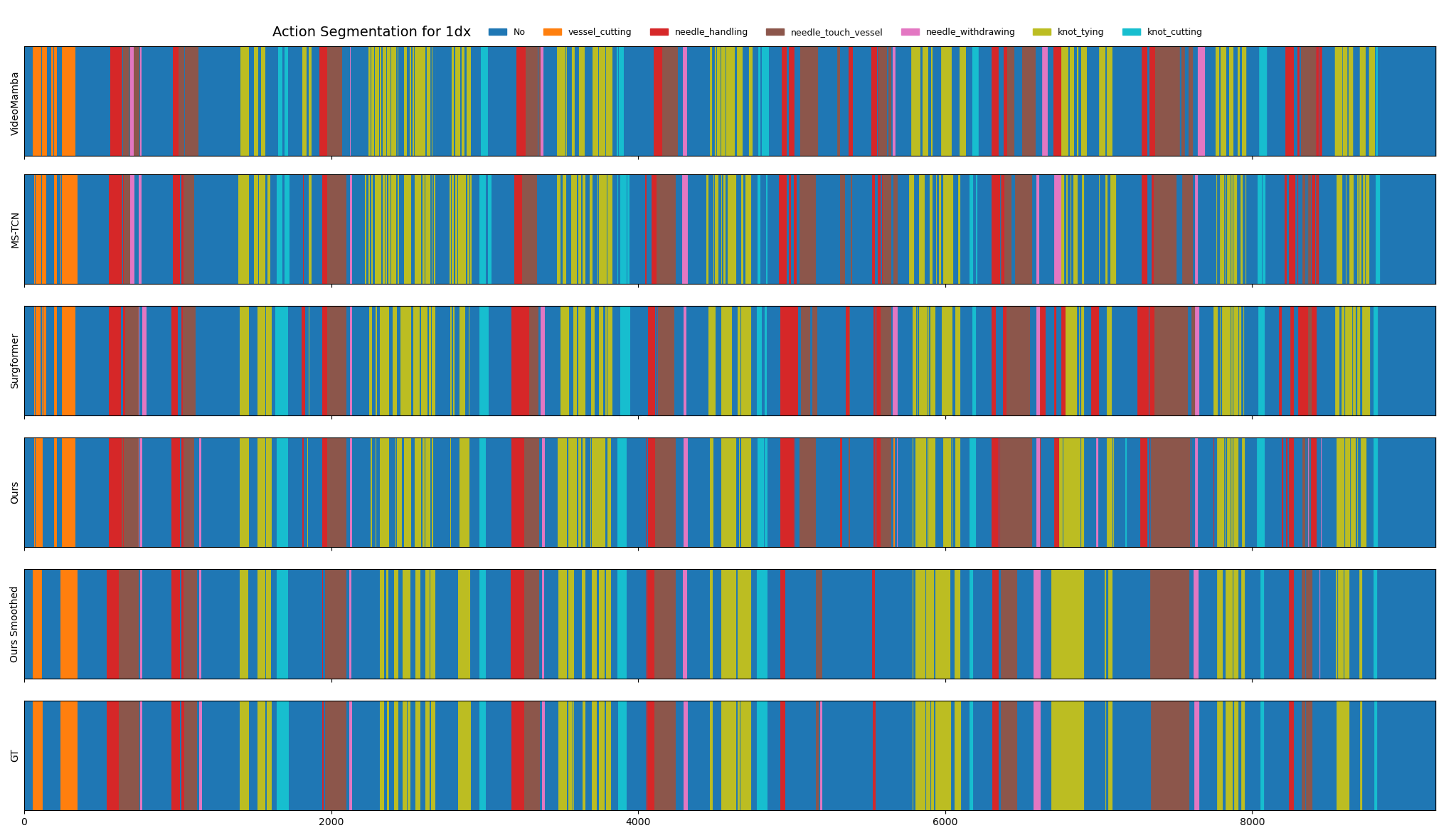}
    \caption{Qualitative action-segmentation comparison with the ground-truth timeline.}
    \label{fig:full_action_seg}
\end{figure}

Both proposed modules improved the baseline (Table~\ref{tab:full_ablation}). Combining HTA and VWSSA increased accuracy by 17.95 percentage points and F1 by 20.55 points. Per-action results in Table~\ref{tab:full_actions} show the strongest F1 scores for vessel cutting, needle--vessel contact, and knot cutting. Needle handling and withdrawal were more difficult because they were brief and visually similar to adjacent actions. The qualitative comparison in Figure~\ref{fig:full_action_seg} shows reduced fragmentation and improved temporal consistency after post-processing.

\par\medskip
\begin{table}[htbp]

\centering
\begin{tabular}{lccccc}
\toprule
Method & Accuracy & Precision & Recall & Jaccard & F1 \\
\midrule
TimeSformer baseline & 69.71 & 57.14 & 55.26 & 43.07 & 62.31 \\
TimeSformer + HTA & 78.27 & 72.45 & 69.55 & 60.20 & 75.45 \\
TimeSformer + VWSSA & 76.50 & 73.29 & 70.42 & 59.36 & 76.84 \\
HTA + VWSSA & \textbf{87.66} & \textbf{83.29} & \textbf{83.99} & \textbf{71.57} & \textbf{82.86} \\
\bottomrule
\end{tabular}
\caption{Ablation study of HTA and VWSSA (\%).}
\label{tab:full_ablation}
\end{table}
\par\medskip

\par\medskip
\begin{table}[htbp]

\centering
\begin{tabular}{lrrrr}
\toprule
Action & Precision & Recall & Jaccard & F1 \\
\midrule
Background & 85.36 & 87.34 & 75.52 & 86.05 \\
Vessel cutting & 86.11 & 89.06 & 77.60 & 87.38 \\
Needle handling & 71.65 & 85.06 & 61.45 & 75.72 \\
Needle touching vessel & 83.18 & 89.42 & 75.78 & 86.01 \\
Needle withdrawal & 84.16 & 67.83 & 62.21 & 74.92 \\
Knot tying & 85.25 & 76.30 & 67.02 & 80.21 \\
Knot cutting & 87.31 & 92.88 & 81.43 & 89.74 \\
\bottomrule
\end{tabular}
\caption{Per-action segmentation performance (\%).}
\label{tab:full_actions}
\end{table}
\par\medskip

\par\medskip
\par\medskip

\subsection{Instrument Tracking and Skill Assessment}
\begin{figure}[htbp]

\centering

    \centering
    \includegraphics[width=0.82\linewidth,height=0.34\textheight,keepaspectratio]{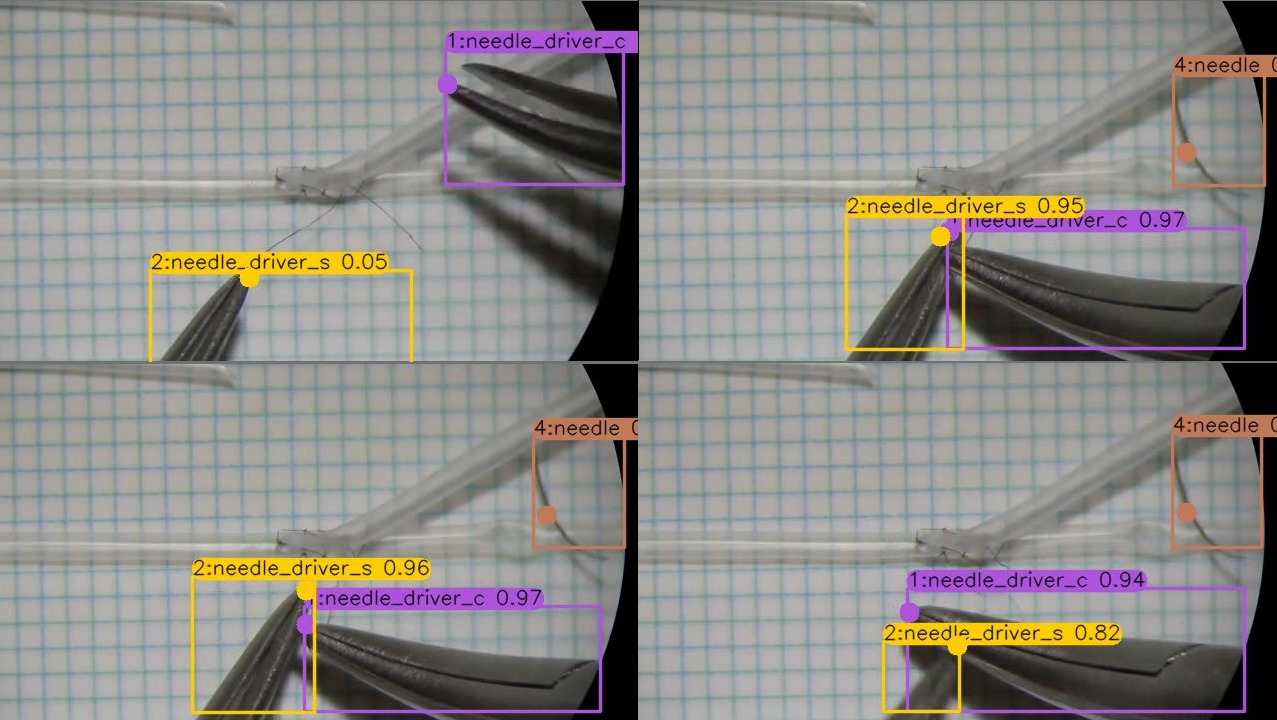}
    \caption{Representative instrument tracking and tip localization.}
    \label{fig:full_tips}
\end{figure}

\begin{figure}[htbp]

\centering
\includegraphics[width=0.78\linewidth,height=0.30\textheight,keepaspectratio]{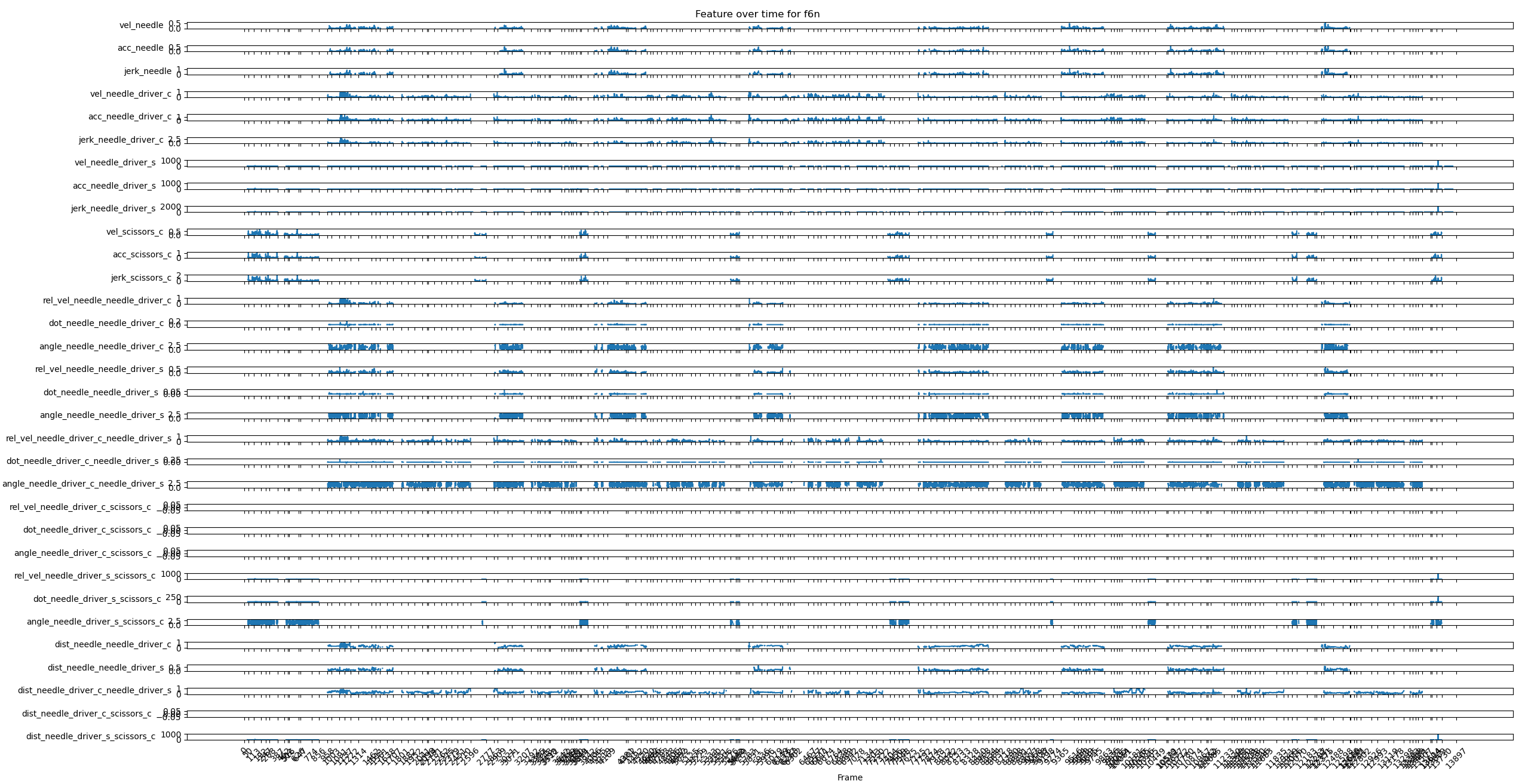}
\caption{Representative kinematic traces used for NOMAT-aligned classification. Peaks mark intervals for source-video inspection; motion features are not direct measurements of tissue force.}
\label{fig:full_features}
\end{figure}

Figure~\ref{fig:full_tips} shows representative detections and localized instrument tips. The resulting trajectories provide the kinematic measures illustrated in Figure~\ref{fig:full_features}. Action duration, repetition, and cumulative time vary substantially across procedures and complement motion features in the downstream classifiers.

\par\medskip
\par\medskip

\par\medskip

The action statistics complement these traces by summarizing the duration, frequency, and cumulative time of each activity. Comparing them with the corresponding video helps distinguish repeated attempts from isolated segmentation fragments.

\begin{figure}[htbp]

\centering
\includegraphics[width=0.78\linewidth,height=0.28\textheight,keepaspectratio]{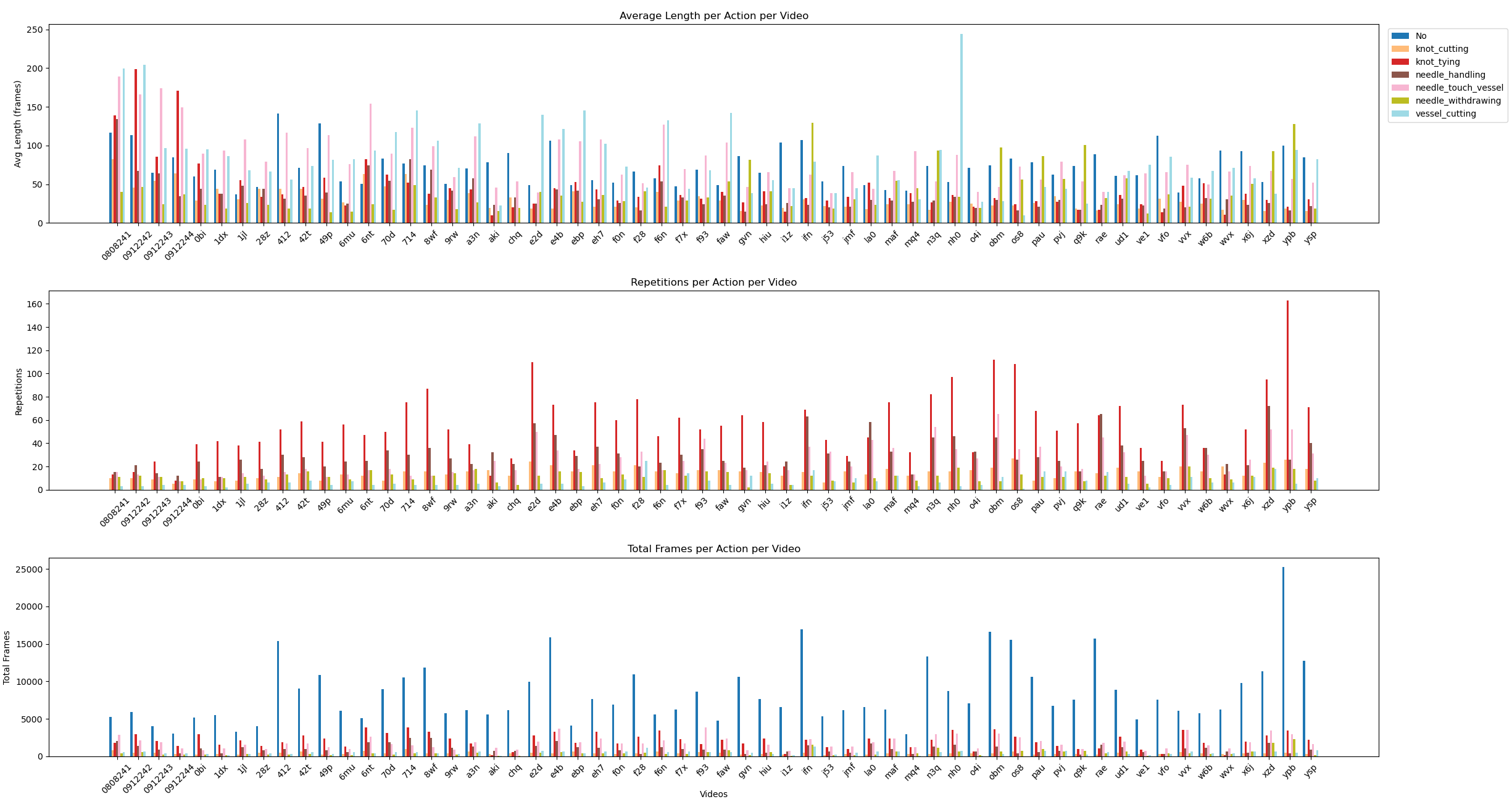}
\caption{Action duration, repetition, and cumulative-time statistics across procedures. Each summary depends on the action timeline and its boundary accuracy.}
\label{fig:full_action_stats}
\end{figure}
\par\medskip

The classifiers achieved 84.8\% accuracy for overall instrument handling, 73.8\% for needle-handling quality, 63.4\% for knot-tying quality, 74.0\% for needle-action efficiency, and 84.0\% for knot-tying efficiency (Table~\ref{tab:full_skill}). Mean accuracy was 76.0\%, and Cohen's $\kappa$ ranged from 0.63 to 0.93.

\par\medskip
\begin{table}[htbp]

\centering
\begin{tabular}{lcc}
\toprule
NOMAT dimension & Accuracy & $\kappa$ \\
\midrule
Instrument handling & 84.8 & 0.88 \\
Needle-handling quality & 73.8 & 0.75 \\
Knot-tying quality & 63.4 & 0.63 \\
Needle-action efficiency & 74.0 & 0.86 \\
Knot-tying efficiency & 84.0 & 0.93 \\
\midrule
Mean & 76.0 & -- \\
\bottomrule
\end{tabular}
\caption{NOMAT-aligned skill-classification results (accuracy in percent).}
\label{tab:full_skill}
\end{table}
\par\medskip

\subsection{Supplementary Detector Validation}
The newer runtime logs report detector precision of 96.3\%, recall of 96.6\%, mAP@0.5 of 97.8\%, and mAP@0.5:0.95 of 92.9\%. These are internal validation statistics from a separate image inventory, not a held-out result for the 72-procedure cohort. The inventory lists 56,981 training, 7,122 validation, and 7,124 test images, whereas the displayed validation run counts 7,121 images. Video-level independence and this one-image discrepancy require reconciliation.

Needle localization is the most difficult class at stricter overlap thresholds, with AP@0.5:0.95 of 78.9\%, compared with 94.6--98.3\% for the instrument categories. Straight scissors has only 59 validation instances. High detection AP does not by itself validate persistent identities, tip localization, or downstream performance assessment. The earlier motion-assessment manuscript reports a different nine-participant cohort with 58 complete recordings; those outcomes are not pooled into the present study.

\subsection{Integrated Interface Demonstration}
The prototype in Figure~\ref{fig:full_ui} demonstrates that outputs from all three modules can be synchronized in one workspace. Selecting a video interval updates the dominant action, visible and tracked instruments, recent motion traces, action statistics, and provisional performance information. The LLM panel can then answer questions using that shared state, allowing a user to move from a summary judgment to the action and motion evidence that supports it. This integration is a functional system contribution, but it is not yet evidence that the LLM responses are consistently correct or improve learning; those questions require a dedicated, task-based evaluation.

\subsection{Interpreting the Feature Visualizations}
The kinematic traces in Figure~\ref{fig:full_features} show how motion changes over a procedure, while action summaries describe duration, repetition, and cumulative time. Their purpose is to connect rubric predictions to observable behavior. A brief speed or acceleration peak identifies an interval to inspect; its meaning depends on whether the participant is handling the needle, contacting the vessel, or tying a knot. An apparent peak can also arise from identity switching, box drift, or a localization error.

Duration measures one action interval, repetition counts its occurrences, and cumulative time sums intervals of a class. Fragmented segmentation can inflate counts and misassign trajectory samples. Conversely, smoothing can erase brief legitimate actions. These relationships explain why downstream classification should be evaluated with both predicted actions and reference actions to quantify propagation of upstream errors. The feature plots are descriptive examples and do not establish a causal relationship between an individual motion measure and surgical competence.

\subsection{User Study Results}
Prior experience was heterogeneous: 10 participants (58.8\%) reported microsurgical training or hands-on experience, 13 (76.5\%) reported microscope or microsurgical exposure, 10 (58.8\%) had performed a practice microanastomosis, and three (17.6\%) had completed a formal microvascular course. At baseline, nine (52.9\%) considered video assessment fair.

Among the three follow-up respondents, task clarity, simulation setting, equipment, value of repetition, perceived improvement, curricular feasibility, willingness to participate again, and support for formal educational use each averaged 4.33 on a five-point scale. Structured video feedback averaged 4.00; reasonableness of blinded scoring and noninterference from recording averaged 3.67; and acceptability of the time burden averaged 3.33. All three reported increased interest in further microsurgical training. Two identified repetition as most helpful. Two identified simulation realism as least helpful and the recording setup as the main interference. Suggestions included aligning camera framing with the microscope view and supporting self-directed recording.

Group-level confidence increased for operating the microscope (2.67 to 3.33), handling instruments (2.44 to 3.33), precise suture placement (2.22 to 3.00), completing a microanastomosis (2.00 to 3.00), and learning through video assessment (2.78 to 3.33). These are descriptive group-level trends rather than paired individual changes.

\par\medskip
\begin{figure}[htbp]
\centering

    \centering
    \begin{minipage}[t]{0.42\textwidth}
        \centering
        \includegraphics[width=\linewidth]{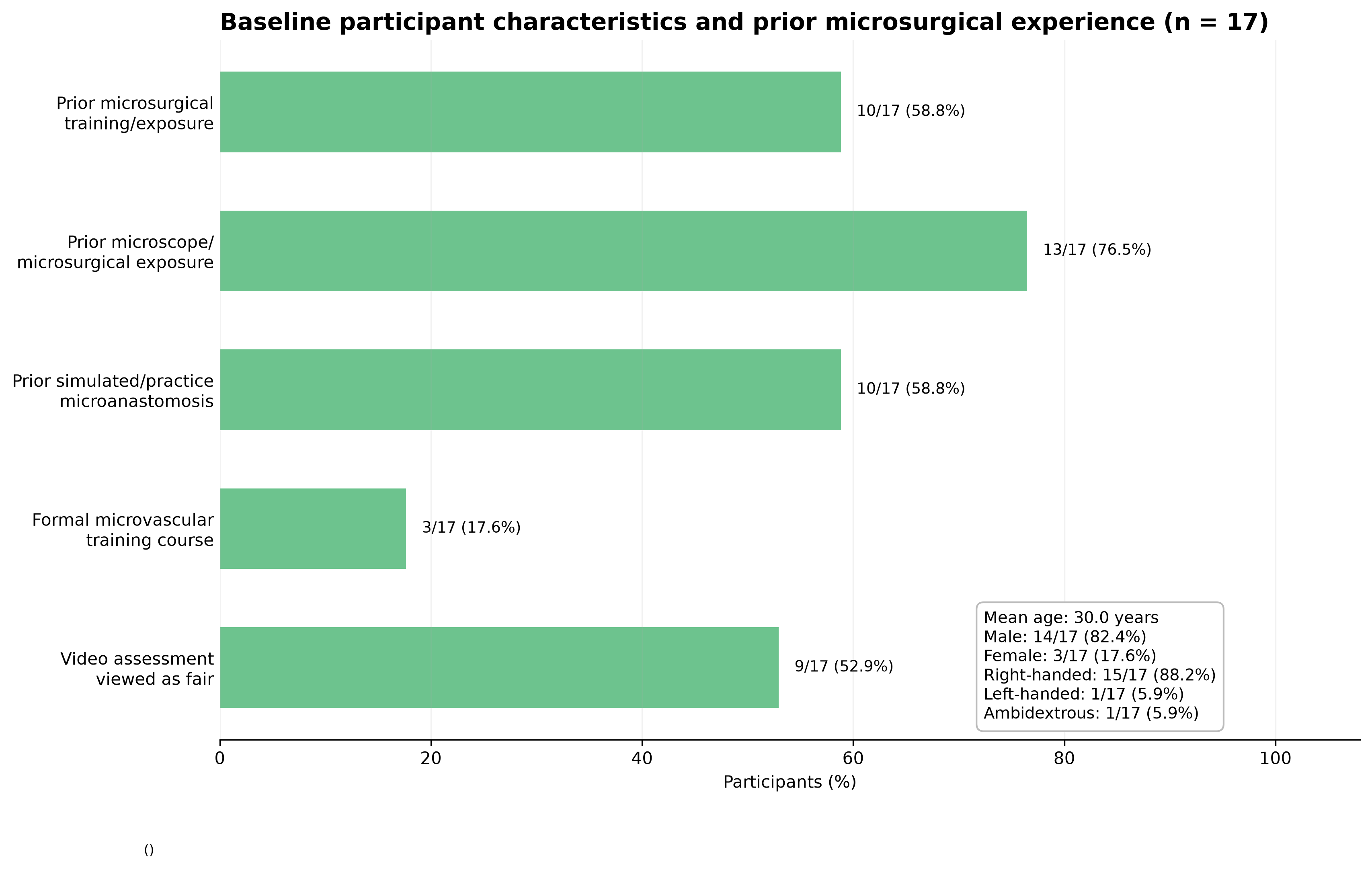}\\[-0.4em]
        \small (a) Baseline characteristics and prior experience.
    \end{minipage}\hfill
    \begin{minipage}[t]{0.42\textwidth}
        \centering
        \includegraphics[width=\linewidth]{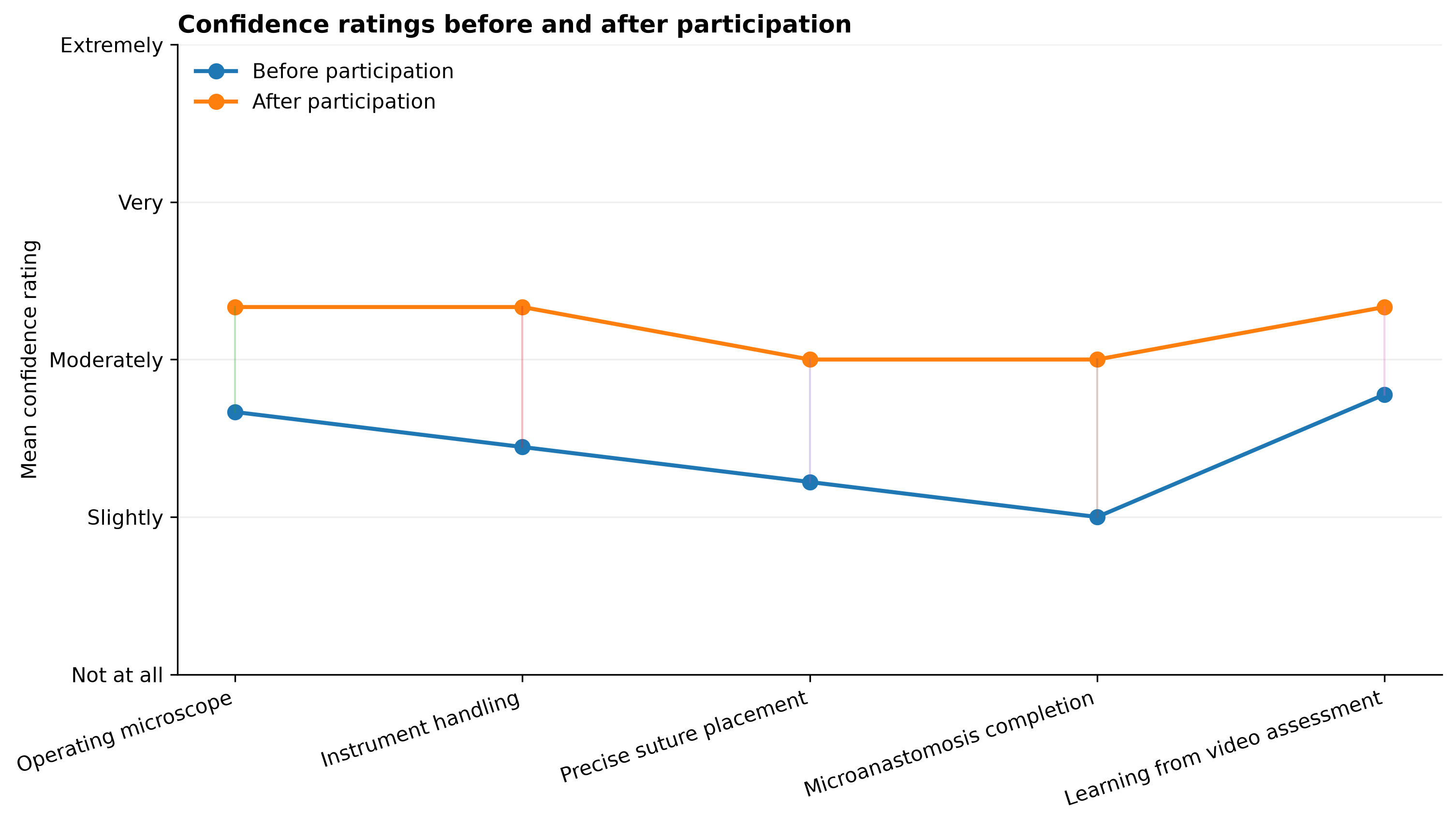}\\[-0.4em]
        \small (b) Group-level confidence before and after participation.
    \end{minipage}\\[0.25em]
    \includegraphics[width=0.60\linewidth,height=0.27\textheight,keepaspectratio]{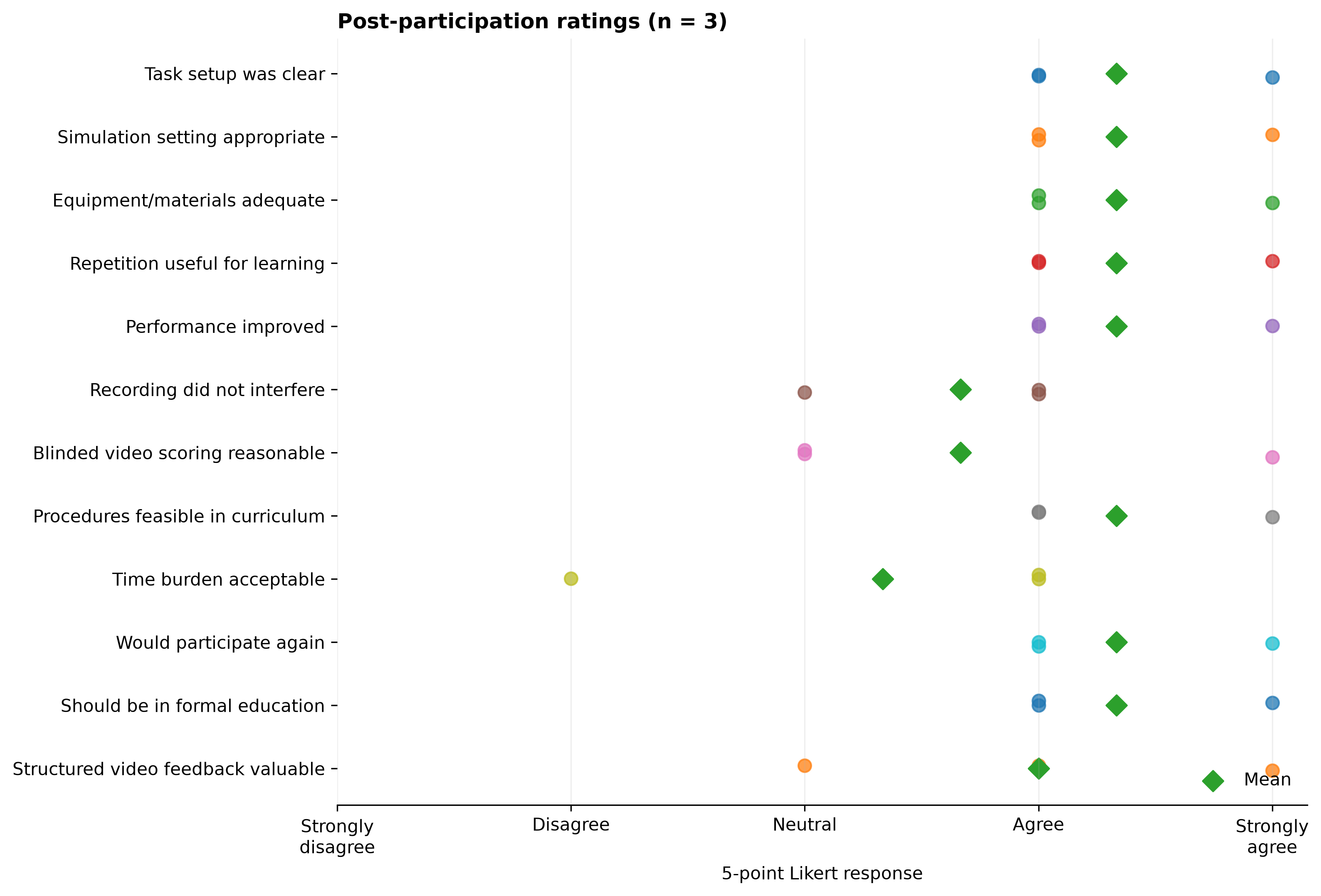}\\[-0.4em]
    \small (c) Follow-up ratings; circles show responses and diamonds show means.
    \caption{Participant survey results. Follow-up values summarize three respondents and are reported descriptively.}
    \label{fig:full_survey}
\end{figure}
\par\medskip

\section{Discussion}
The principal contribution is an integrated platform that connects three distinct capabilities needed for useful video-based training support. Action segmentation supplies temporal structure; object detection and tracking quantify how tools move within that structure; supervised classification relates the resulting kinematic and action statistics to expert performance dimensions; and the grounded LLM makes the shared evidence accessible through questions in the same interface. This organization is more useful than an isolated segmentation label or procedure-level score because users can move between an assessment, its measurable basis, and the source video.

The action-segmentation model remains the main methodological contribution. Its multiscale temporal and variance-weighted spatial attention improves recognition of subtle, variable-duration microsurgical actions, while workflow-aware refinement produces a coherent timeline for downstream analysis. The downstream performance results show why this accuracy matters at the system level: action boundaries determine which instrument trajectories, motion features, and temporal statistics are assigned to each activity and subsequently used by the supervised classifiers.

Several limitations remain. The dataset is small, only two manually annotated procedures formed the segmentation test set, and all recordings were collected under controlled simulation conditions. Generalization must be evaluated across institutions, microscopes, cameras, instruments, lighting, participant backgrounds, and handedness. The skill-classification split should also be replaced by larger participant-independent external evaluations with confidence intervals and calibration analysis. The LLM module requires a grounded question-answering benchmark covering answer correctness, evidence attribution, unsupported claims, abstention, and human factors. Finally, the follow-up survey had only three respondents and did not link individual baseline and follow-up responses. It indicates acceptability but does not establish educational effectiveness.

Future prospective studies should compare AI-assisted review with conventional feedback and measure processing failure, abstention, faculty review time, feedback timeliness, skill retention, and transfer to new tasks. Predictions should remain provisional and should not be used for credentialing or punitive monitoring without separate validation, governance, and consent.

\section{Conclusion}
We presented an integrated microanastomosis training platform in which a proposed action-segmentation model organizes video into task-relevant intervals, instrument detection and tracking produce action-conditioned kinematic evidence, supervised classifiers estimate five NOMAT-aligned performance dimensions, and a grounded LLM supports scene- and performance-related interaction through a unified interface. Across two sites, 17 participants completed 72 procedures comprising 576 suture placements. Segmentation reached 93.62\% accuracy after refinement, and the performance classifiers achieved 76.0\% mean accuracy. These results establish the technical basis of the integrated system while motivating dedicated evaluation of language grounding, cross-site generalization, user interaction, and educational benefit.

\bibliography{aaai2027}

@article{meara2015global,
	title={Global Surgery 2030: evidence and solutions for achieving health, welfare, and economic development},
	author={Meara, John G and Leather, Andrew JM and Hagander, Lars and Alkire, Blake C and Alonso, Nivaldo and Ameh, Emmanuel A and Bickler, Stephen W and Conteh, Lesong and Dare, Anna J and Davies, Justine and others},
	journal={The lancet},
	volume={386},
	number={9993},
	pages={569--624},
	year={2015},
	publisher={Elsevier}
}

@inproceedings{redmon2016you,
	title={You only look once: Unified, real-time object detection},
	author={Redmon, J},
	booktitle={Proceedings of the IEEE conference on computer vision and pattern recognition},
	year={2016}
}

@inproceedings{wojke2017simple,
	title={Simple online and realtime tracking with a deep association metric},
	author={Wojke, Nicolai and Bewley, Alex and Paulus, Dietrich},
	booktitle={2017 IEEE international conference on image processing (ICIP)},
	pages={3645--3649},
	year={2017},
	organization={IEEE}
}

@article{aoun2015pilot,
	title={A pilot study to assess the construct and face validity of the Northwestern Objective Microanastomosis Assessment Tool},
	author={Aoun, Salah G and El Ahmadieh, Tarek Y and El Tecle, Najib E and Daou, Marc R and Adel, Joseph G and Park, Christine S and Batjer, H Hunt and Bendok, Bernard R},
	journal={Journal of neurosurgery},
	volume={123},
	number={1},
	pages={103--109},
	year={2015},
	publisher={American Association of Neurological Surgeons}
}

@article{martin1997objective,
	title={Objective structured assessment of technical skill (OSATS) for surgical residents},
	author={Martin, JA and Regehr, Glenn and Reznick, Richard and Macrae, Helen and Murnaghan, John and Hutchison, Carol and Brown, M},
	journal={British journal of surgery},
	volume={84},
	number={2},
	pages={273--278},
	year={1997},
	publisher={Wiley Online Library}
}

@article{zia2016automated,
	title={Automated video-based assessment of surgical skills for training and evaluation in medical schools},
	author={Zia, Aneeq and Sharma, Yachna and Bettadapura, Vinay and Sarin, Eric L and Ploetz, Thomas and Clements, Mark A and Essa, Irfan},
	journal={International journal of computer assisted radiology and surgery},
	volume={11},
	pages={1623--1636},
	year={2016},
	publisher={Springer}
}

@inproceedings{meng2023automatic,
	title={An Automatic Grading System for Neonatal Endotracheal Intubation with Multi-Task Convolutional Neural Network},
	author={Meng, Yan and Hahn, James K},
	booktitle={2023 IEEE EMBS International Conference on Biomedical and Health Informatics (BHI)},
	pages={1--4},
	year={2023},
	organization={IEEE}
}

@article{funke2019video,
	title={Video-based surgical skill assessment using 3D convolutional neural networks},
	author={Funke, Isabel and Mees, S{\"o}ren Torge and Weitz, J{\"u}rgen and Speidel, Stefanie},
	journal={International journal of computer assisted radiology and surgery},
	volume={14},
	pages={1217--1225},
	year={2019},
	publisher={Springer}
}

@article{konstantinov2021interpretable,
	title={Interpretable machine learning with an ensemble of gradient boosting machines},
	author={Konstantinov, Andrei V and Utkin, Lev V},
	journal={Knowledge-Based Systems},
	volume={222},
	pages={106993},
	year={2021},
	publisher={Elsevier}
}

@article{regehr1998comparing,
	title={Comparing the psychometric properties of checklists and global rating scales for assessing performance on an OSCE-format examination},
	author={Regehr, Glenn and MacRae, Helen and Reznick, Richard K and Szalay, David},
	journal={Academic Medicine},
	volume={73},
	number={9},
	pages={993--7},
	year={1998},
	publisher={LWW}
}

@article{lavanchy2021automation,
	title={Automation of surgical skill assessment using a three-stage machine learning algorithm},
	author={Lavanchy, Jo{\"e}l L and Zindel, Joel and Kirtac, Kadir and Twick, Isabell and Hosgor, Enes and Candinas, Daniel and Beldi, Guido},
	journal={Scientific reports},
	volume={11},
	number={1},
	pages={5197},
	year={2021},
	publisher={Nature Publishing Group UK London}
}

@article{qiu2019real,
	title={Real-time surgical instrument tracking in robot-assisted surgery using multi-domain convolutional neural network},
	author={Qiu, Liang and Li, Changsheng and Ren, Hongliang},
	journal={Healthcare technology letters},
	volume={6},
	number={6},
	pages={159--164},
	year={2019},
	publisher={Wiley Online Library}
}

@inproceedings{czempiel2020tecno,
	title={Tecno: Surgical phase recognition with multi-stage temporal convolutional networks},
	author={Czempiel, Tobias and Paschali, Magdalini and Keicher, Matthias and Simson, Walter and Feussner, Hubertus and Kim, Seong Tae and Navab, Nassir},
	booktitle={Medical Image Computing and Computer Assisted Intervention--MICCAI 2020: 23rd International Conference, Lima, Peru, October 4--8, 2020, Proceedings, Part III 23},
	pages={343--352},
	year={2020},
	organization={Springer}
}

@inproceedings{farha2019ms,
	title={Ms-tcn: Multi-stage temporal convolutional network for action segmentation},
	author={Farha, Yazan Abu and Gall, Jurgen},
	booktitle={Proceedings of the IEEE/CVF conference on computer vision and pattern recognition},
	pages={3575--3584},
	year={2019}
}

@inproceedings{bertasius2021space,
	title={Is space-time attention all you need for video understanding?},
	author={Bertasius, Gedas and Wang, Heng and Torresani, Lorenzo},
	booktitle={ICML},
	volume={2},
	pages={4},
	year={2021}
}

@inproceedings{zhang2022actionformer,
	title={Actionformer: Localizing moments of actions with transformers},
	author={Zhang, Chen-Lin and Wu, Jianxin and Li, Yin},
	booktitle={European Conference on Computer Vision},
	pages={492--510},
	year={2022},
	organization={Springer}
}

@inproceedings{yang2024surgformer,
	title={Surgformer: Surgical transformer with hierarchical temporal attention for surgical phase recognition},
	author={Yang, Shu and Luo, Luyang and Wang, Qiong and Chen, Hao},
	booktitle={International Conference on Medical Image Computing and Computer-Assisted Intervention},
	pages={606--616},
	year={2024},
	organization={Springer}
}

@inproceedings{wang2018satr,
	title={SATR-DL: improving surgical skill assessment and task recognition in robot-assisted surgery with deep neural networks},
	author={Wang, Ziheng and Fey, Ann Majewicz},
	booktitle={2018 40th Annual International Conference of the IEEE Engineering in Medicine and Biology Society (EMBC)},
	pages={1793--1796},
	year={2018},
	organization={IEEE}
}

@inproceedings{li2022surgical,
	title={Surgical skill assessment via video semantic aggregation},
	author={Li, Zhenqiang and Gu, Lin and Wang, Weimin and Nakamura, Ryosuke and Sato, Yoichi},
	booktitle={International Conference on Medical Image Computing and Computer-Assisted Intervention},
	pages={410--420},
	year={2022},
	organization={Springer}
}

@article{kitaguchi2021development,
	title={Development and validation of a 3-dimensional convolutional neural network for automatic surgical skill assessment based on spatiotemporal video analysis},
	author={Kitaguchi, Daichi and Takeshita, Nobuyoshi and Matsuzaki, Hiroki and Igaki, Takahiro and Hasegawa, Hiro and Ito, Masaaki},
	journal={JAMA network open},
	volume={4},
	number={8},
	pages={e2120786--e2120786},
	year={2021},
	publisher={American Medical Association}
}

@article{goldbraikh2022video,
	title={Video-based fully automatic assessment of open surgery suturing skills},
	author={Goldbraikh, Adam and D’Angelo, Anne-Lise and Pugh, Carla M and Laufer, Shlomi},
	journal={International Journal of Computer Assisted Radiology and Surgery},
	volume={17},
	number={3},
	pages={437--448},
	year={2022},
	publisher={Springer}
}

@article{hung2023capturing,
	title={Capturing fine-grained details for video-based automation of suturing skills assessment},
	author={Hung, Andrew J and Bao, Richard and Sunmola, Idris O and Huang, De-An and Nguyen, Jessica H and Anandkumar, Anima},
	journal={International journal of computer assisted radiology and surgery},
	volume={18},
	number={3},
	pages={545--552},
	year={2023},
	publisher={Springer}
}

@article{zhang2023surgical,
	title={Surgical workflow recognition with temporal convolution and transformer for action segmentation},
	author={Zhang, Bokai and Goel, Bharti and Sarhan, Mohammad Hasan and Goel, Varun Kejriwal and Abukhalil, Rami and Kalesan, Bindu and Stottler, Natalie and Petculescu, Svetlana},
	journal={International Journal of Computer Assisted Radiology and Surgery},
	volume={18},
	number={4},
	pages={785--794},
	year={2023},
	publisher={Springer}
}

@article{yi2021asformer,
	title={Asformer: Transformer for action segmentation},
	author={Yi, Fangqiu and Wen, Hongyu and Jiang, Tingting},
	journal={arXiv preprint arXiv:2110.08568},
	year={2021}
}

@article{kay2017kinetics,
	title={The kinetics human action video dataset},
	author={Kay, Will and Carreira, Joao and Simonyan, Karen and Zhang, Brian and Hillier, Chloe and Vijayanarasimhan, Sudheendra and Viola, Fabio and Green, Tim and Back, Trevor and Natsev, Paul and others},
	journal={arXiv preprint arXiv:1705.06950},
	year={2017}
}

@article{yanik2023video,
	title={Video-based formative and summative assessment of surgical tasks using deep learning},
	author={Yanik, Erim and Kruger, Uwe and Intes, Xavier and Rahul, Rahul and De, Suvranu},
	journal={Scientific Reports},
	volume={13},
	number={1},
	pages={1038},
	year={2023},
	publisher={Nature Publishing Group UK London}
}

@inproceedings{meng2025ai,
	title={An AI Framework for Microanastomosis Motion Assessment},
	author={Meng, Yan and Torres-Rodr{\'\i}guez, Eduardo J and Altshuler, Marcelle and Gowda, Nishanth and Naeem, Arhum and Yilmaz, Recai and Arnaout, Omar and Donoho, Daniel A},
	booktitle={2025 12th International IEEE/EMBS Conference on Neural Engineering (NER)},
	pages={132--137},
	year={2025},
	organization={IEEE}
}
\end{document}